\documentclass[11pt,a4paper]{article}
\usepackage[hyperref]{acl2021}
\usepackage{times}
\usepackage{latexsym}

\usepackage{microtype}

\usepackage{graphicx}
\usepackage{booktabs}
\usepackage{colortbl}
\usepackage{tablefootnote}
\usepackage{xcolor}
\usepackage{bm}
\usepackage{amsmath}
\usepackage{caption}
\usepackage{subcaption}
\usepackage{soul}
\usepackage{multirow}
\usepackage{supertabular}
\usepackage{tabularx}
\usepackage{arydshln}
\usepackage[outline]{contour}

\usepackage{placeins}
\usepackage{float}

\newcolumntype{L}[1]{>{\raggedright\let\newline\\\arraybackslash\hspace{0pt}}m{#1}}

\newcommand{\hlc}[2][yellow]{{%
    \colorlet{foo}{#1}%
    \sethlcolor{foo}\hl{#2}}%
}

\PassOptionsToPackage{dvipsnames}{xcolor}

\usepackage[group-separator={,}]{siunitx}
\aclfinalcopy 
\def\aclpaperid{3635} 

\title{Controllable Open-ended Question Generation \\with A New Question Type Ontology}

\author{Shuyang Cao \and Lu Wang \\
  Computer Science and Engineering \\
  University of Michigan \\
  Ann Arbor, MI \\
  \texttt{\{caoshuy, wangluxy\}@umich.edu} \\}

\date{}

\begin{document}
\maketitle

\begin{abstract}
We investigate the less-explored task of generating \textit{open-ended questions} that are typically answered by multiple sentences.
We first define a \textit{new question type ontology} which differentiates the nuanced nature of questions better than widely used question words. A new dataset with $4,959$ questions is labeled based on the new ontology. 
We then propose a novel question type-aware question generation framework, augmented by a semantic graph representation, to jointly predict question focuses and produce the question. 
Based on this framework, we further use both exemplars and automatically generated templates to improve controllability and diversity. 
Experiments on two newly collected large-scale datasets show that our model improves question quality over competitive comparisons based on automatic metrics. Human judges also rate our model outputs highly in answerability, coverage of scope, and overall quality. Finally, our model variants with templates can produce questions with enhanced controllability and diversity. 

\end{abstract}
\section{Introduction}

\begin{figure}[t]
    \fontsize{9}{11}\selectfont
    \setlength{\tabcolsep}{0pt}
    \hspace{-1mm}
    \begin{tabular}{p{0.48\textwidth}}
        \toprule
        \textbf{Input}: It's a difficult task to undertake. Teenagers tend to identify gangs with ``fitting'' in. Peer pressure plays a large part in it and sometimes teenagers have problems with their own identity being part of a gang deals with those issues. It also provides a little bit of respect on the street ... \\ 
        \midrule
        \textbf{\textsc{BART Sampling}}: \\
        - How do you stop a teen from joining a gang? (\textsc{Procedural}) \\
        - How do you get teenagers to stop being in gangs? \\ (\textsc{Procedural}) \\
        - How do you get teens out of gangs? (\textsc{Procedural}) \\
        \textbf{\textsc{BART + QWord}}: \\
        - \textcolor{red!70!black}{\textbf{How}} do you get a teenager out of a gang? (\textsc{Procedural}) \\
        - \textcolor{red!70!black}{\textbf{What}} is the best way to get teenagers out of gangs? \\ (\textsc{Procedural}) \\
        - \textcolor{red!70!black}{\textbf{Why}} do teenagers join gangs? (\textsc{Cause}) \\
        \textbf{\textsc{TplGen}}: \\
        - How do I get \texttt{[NP]} to quit being in \texttt{[NP]}? \contour{black}{$\Rightarrow$} \textcolor{blue!60!black}{\textbf{How do I get}} my son \textcolor{blue!60!black}{\textbf{to quit being in}} a gang? (\textsc{Procedural}) \\
        - What are \texttt{[NP]}? \contour{black}{$\Rightarrow$} \textcolor{blue!60!black}{\textbf{What are}} some programs for teenagers involved in gangs? (\textsc{Example}) \\
        - Why do \texttt{[NP]} \texttt{[V]} \texttt{[NP]}? \contour{black}{$\Rightarrow$} \textcolor{blue!60!black}{\textbf{Why do}} teenagers identify gangs? (\textsc{Cause}) \\
        \bottomrule
    \end{tabular}
    \vspace{-2mm}
    \caption{
    Open-ended questions generated by different models after reading the same input: (1) BART decoded with nucleus sampling, (2) BART that considers different question words, and (3) our type-aware generator \textsc{TplGen}, that predicts focuses and operates with generated templates (to the left of the arrows).
    Questions generated by our model have diverse \textsc{Type}s. 
    }
    \vspace{-3mm}
    \label{fig:intro}
\end{figure}

Question-asking has long served as an effective instrument for knowledge learning~\cite{10.2307/1169962,tobin1990research} and assessing learning progress~\cite{holme2003assessment,downing2009assessment,livingston2009constructed}. 
Compared to the widely studied task of generating factoid questions that inquire about ``one bit" of information~\cite{du-etal-2017-learning,duan-etal-2017-question,li-etal-2019-improving-question}, this work is interested in \textit{generating open-ended questions} that require deep comprehension and long-form answers~\cite{labutov-etal-2015-deep}. Such open-ended questions are valuable in education, e.g., to facilitate complex knowledge acquisition~\cite{lai2017race} and nurture reasoning skills~\cite{shapley2000line}, as well as in other applications like improving search engines~\cite{han2019inferring} and building open-domain dialogue systems~\cite{shum2018eliza}.

Significant progress has been made in generating factoid questions~\cite{zhang-bansal-2019-addressing,zhou-etal-2019-question,su-etal-2020-multi}, yet new challenges need to be addressed for open-ended questions. 
First, \textit{specifying the question type} is crucial for constructing meaningful questions~\cite{graesser1992mechanisms}. Question words such as ``why" and ``when" are generally seen as being indicative of types~\cite{zhou-etal-2019-question}, but they underspecify the conceptual content of questions~\cite{olney2012question}.
Using Figure~\ref{fig:intro} as an example, different question words, i.e., both ``how" and ``what", can be used for inquiring about procedures. It thus calls for a new question type ontology that can \textit{precisely capture the conceptual nature of questions}. 
Second, constructing questions from a text with multiple sentences needs to focus on its \textit{central concepts or phenomena that necessitate extensive descriptions}. New representations are needed to capture such content as question focus(es), to go beyond existing methods that rely on entities and their neighboring words~\cite{du-etal-2017-learning,sun-etal-2018-answer} even though they are effective for generating factoid questions. 
Third, encouraging the \textit{diversity} of generated questions~\cite{sultan-etal-2020-importance,wang-etal-2020-diversify} is less explored but critical for real world applications, e.g., various questions should be proposed to gauge how well students grasp the knowledge of complex subjects.

In this work, we aim to address the challenges of generating open-ended questions from input consisting of multiple sentences.
We first introduce \textbf{a new question type ontology}, drawn upon researches in cognitive science and psychology~\cite{graesser1992mechanisms}, to capture deeper levels of cognition, such as causal reasoning and judgments. Based on the new ontology, we collect and annotate a dataset of $4{,}959$ questions to benefit research in both question generation and answering.\footnote{Our data and code are available at: \url{https://shuyangcao.github.io/projects/ontology_open_ended_question}.} 

We then design \textbf{a type-aware framework} to jointly predict question focuses ({what to ask about}) and generate questions ({how to ask it}). Different from pipeline-based approaches (e.g., \newcite{sun-etal-2018-answer}), our framework is built on large pre-trained BART~\cite{lewis-etal-2020-bart}, and uses shared representations to jointly conduct question focus prediction and question generation while learning task-specific knowledge. 
It is further augmented by a semantic graph that leverages both semantic roles and dependency relations, facilitating long text comprehension to pinpoint salient concepts. 

Moreover, to achieve the goal of producing various types of questions from the same input, we investigate two model variants that \textbf{use templates to improve controllability and generation diversity}: one using pre-identified exemplars, the other employing generated templates to guide question writing, with sample outputs displayed in Figure~\ref{fig:intro}.

For experiments, we collect two new large-scale datasets consisting of open-ended questions with answers from (1) Yahoo Answers\footnote{\url{https://answers.yahoo.com/}} L6 dataset and (2) popular question-asking communities on Reddit\footnote{\url{https://www.reddit.com/}}, consisting of $291$K and $720$K question-answer pairs, respectively. Compared to existing popular QA datasets, such as SQuAD~\cite{rajpurkar-etal-2016-squad} and MS MARCO~\cite{bajaj2016ms}), questions in our datasets ask about complex phenomena and perplexing social issues that seek solutions expressed in a long form. 
Automatic metrics show that our type-aware question generation model outperforms competitive comparisons, highlighting the effectiveness of semantic graph-augmented representation and joint modeling of focus prediction and question generation. 
Human judges also confirm that questions generated by our model have better overall quality. 
Adding templates further promotes question diversity, as evaluated by both automatic evaluation and human assessment.

\section{Related Work}

Question generation has long been studied to reduce human efforts in constructing questions for knowledge learning evaluation~\cite{mitkov-ha-2003-computer,brown2005automatic}. 
Early work relies on syntactic transformation to convert declarative sentences to questions~\cite{heilman-smith-2010-good,chali2015towards}. 
Recent advancements rely on sequence-to-sequence models to generate a question from a given sentence or paragraph by considering the focus, type, and general-specific relations of questions~\cite{sun-etal-2018-answer,zhou-etal-2019-question,krishna-iyyer-2019-generating}. 
In particular, question likelihoods and rewards are designed to steer them toward being addressed by the given answers~\cite{zhou-etal-2019-multi,zhang-bansal-2019-addressing}. 
Attempts are also made toward creating complex questions that require multi-hop reasoning over the given text, and graph-based representations have been an enabling tool to facilitate the access to both entities and relations~\cite{pan-etal-2020-semantic, su-etal-2020-multi}. While our model also enhances the input with a semantic graph, it boasts a richer representation by including both dependency and semantic relations, with predicted question focuses highlighted via extra node embeddings. Moreover, we create a separate layer of cross attentions that is dedicated to the semantic graph, while prior work uses the same set of attentions to attend to the concatenated text and graph representations. 

Given the data-driven nature of question generation and answering tasks, recent studies take advantage of the availability of large-scale QA datasets, such as SQuAD~\cite{rajpurkar-etal-2016-squad}, MS MARCO~\cite{bajaj2016ms}, HotpotQA~\cite{yang-etal-2018-hotpotqa}, DROP~\cite{dua2019drop}, inter alia. 
These corpora mainly contain factoid questions, while our newly collected datasets are not only larger in size but also comprise significantly more open-ended questions for querying reasons and procedures. 
A dataset closer to ours is ELI5~\cite{fan-etal-2019-eli5}, which also obtains open-ended question-answer pairs from Reddit, while one of our datasets includes more Reddit communities and thus covers a wider range of topics.

Our work is more inline with generating deeper questions with responses that span over multiple sentences, where manually constructed templates are found effective~\cite{olney2012question}. For example, \newcite{labutov-etal-2015-deep} use crowdsourcing to collect question templates based on an ontology derived from Wikipedia and Freebase topics. Different from the topic-based ontology, our question types are more aligned with cognitive levels. Moreover, our templates are automatically learned from training data. 
Recent work~\cite{rao-daume-iii-2018-learning,rao-daume-iii-2019-answer} focuses on asking clarification questions based on both retrieval and generation models. As there has been no suitable framework for diverse types of questions, this work aims to fill the gap by introducing type-aware generation models which optionally leverage question templates for better controllability.

Generating diverse questions is much less studied, with existing approaches mainly focusing on entity replacement~\cite{cho-etal-2019-mixture}, sampling decoding~\cite{sultan-etal-2020-importance,wang-etal-2020-diversify}, and post-filtering~\cite{liu2020asking}. However, the produced diversity is driven by word choice and syntax variation, with little ability to control on question types, which is the focus of this work.

\section{Data Collection and Question Type Annotation}

\subsection{Open-ended Question Datasets}
To collect open-ended questions, we resort to online forums with active question-asking discussions. Concretely, we gather and clean \textit{question-answer pairs} from Reddit and Yahoo Answers, to train generators that construct questions by taking the corresponding answer as input. 

We choose five popular \textbf{Reddit} communities: \texttt{r/AskHistorians}, \texttt{r/Ask\_Politics}, \texttt{r/askscience}, \texttt{r/explainlikeimfive}, and \texttt{r/AskReddit}, where open-ended questions are actively asked. The original posts (OPs) are extracted, with their titles becoming questions. We also keep the best answer with the highest karma (i.e., upvotes minus downvotes) if it is greater than 1. A second dataset with question-answer pairs is collected from the \textbf{Yahoo} Answers L6 corpus\footnote{\url{https://webscope.sandbox.yahoo.com/}}, which covers a broader range of topics than the Reddit data. For each question, the best answer is rated by the user who raises the question. 

\smallskip
\noindent \textbf{Preprocessing.} 
To ensure both questions and answers are well-formed, human inspection is conducted in multiple iterations to design rules to filter out improper samples. For instance, we discard samples whose answers have less than $15$ content words to avoid the inclusion of factoid question. More details are provided in Table~\ref{tab:data_clean_rule} in Appendix~\ref{appendix:data_collection}. 
Ultimately, $719{,}988$ question-answer pairs are kept for Reddit, and $290{,}611$ for Yahoo. 
Each dataset is then divided into train, validation and test sets with a $90\%$/$5\%$/$5\%$ split. The average lengths of questions and answers are $14.5$ and $117.8$ for Reddit, and $12.2$ and $123.6$ for Yahoo.

\subsection{Question Type Ontology and Annotation}
\label{subsec:question_type_ontology}

\begin{table}[t]
    \centering
    \fontsize{8}{10}\selectfont
    \setlength{\tabcolsep}{0.3mm}
    \begin{tabular}{lp{0.35\textwidth}}
    \toprule
        \textbf{Question Type} & \textbf{Description} (asking for...) \\
        \midrule
        \textsc{Verification} & the truthfulness of an event or a concept. \\
        \midrule
        \textsc{Disjunctive} & the true one given multiple events or concepts, where comparison among options is not needed. \\
        \midrule
        \textsc{Concept} & a definition of an event or a concept. \\
        \midrule
        \textsc{Extent} & the extent or quantity of an event or a concept. \\
        \midrule
        \textsc{Example} & example(s) or instance(s) of an event or a concept. \\
        \midrule
        \textsc{Comparison} & comparison among multiple events or concepts. \\
        \midrule
        \textsc{Cause} &  the cause or reason for an event or a concept. \\
        \midrule
        \textsc{Consequence} & the consequences or results of an event. \\
        \midrule
        \textsc{Procedural} & the procedures, tools, or methods by which a certain outcome is achieved. \\
        \midrule
        \textsc{Judgmental} & the opinions of the answerer’s own. \\
        \bottomrule
    \end{tabular}
    \caption{
    Our new question type ontology, which is adopted and modified from \newcite{olney2012question}. Types are sorted by levels of cognition (lower to higher).}
    \label{tab:type_def}
\end{table}

Our question type ontology is adopted and modified from \citet{olney2012question}, where $18$ categories are originally proposed for knowledge learning assessment. 
We recruited $6$ native English speakers for three rounds of question type annotation. 
Based on the annotators' feedback after each round, we refine the definitions, merge ambiguous types, and delete inapplicable categories. For example, an initial \textsc{Expectation} type is merged into \textsc{Cause} due to their similarities in seeking causality. 
Finally, $10$ types are preserved (Table~\ref{tab:type_def}). As can be seen, our ontology is designed to better capture the nature of questions than question words. 

\smallskip
\noindent \textbf{Annotating Questions with Types.} 
After the annotation guideline is finalized, we ask the same set of annotators to label $5{,}000$ ($2 \times 2{,}500$) randomly sampled questions from both Reddit and Yahoo's training sets. Each question is labeled by two annotators, with disagreements resolved through discussions. After removing samples without consensus, the final dataset consists of $4,959$ questions. \textsc{Example} questions are most prevalent, comprising $23.4\%$ of samples, while only $2.6\%$ are \textsc{Consequence} questions. 
A Krippendorff's $\alpha$ of $0.67$ is obtained for all samples, indicating a reasonable agreement level.
The annotation guideline and examples for each question type are shown in Table~\ref{fig:ques_type_anno_guideline} in Appendix~\ref{appendix:data_collection}.

\smallskip
\noindent \textbf{Training Question Type Classifiers.} 
Since our type-aware question generation model requires a specified type as input, here we describe how to build two question type classifiers: 
(1) $\gamma_q$, that labels a type by \textit{reading the question} and is used to provide question type labels during \textit{training}; 
(2) $\gamma_a$, that predicts a type for use by \textit{taking the answer as input} and is used during \textit{test}. 

Both classifiers are based on RoBERTa~\cite{liu2019roberta}, where a prediction layer is built on top of the contextual representation of the \texttt{[BOS]} token to output question type probabilities.
$\gamma_q$ achieves a macro F1 score of $0.80$ on a reserved test set, with data splits detailed in Appendix~\ref{appendix:type_classifier}. 
To train $\gamma_a$, in addition to the annotated questions, we run $\gamma_q$ on unlabeled questions in Reddit and Yahoo and include samples whose type prediction confidence score is above $0.9$. 
We train one $\gamma_a$ for each dataset. $\gamma_a$ obtains macro F1 scores of $0.48$ and $0.46$ on the same reserved test set over all types after training on Yahoo and Reddit, respectively.

After running $\gamma_q$ on both datasets, we find that Reddit has significantly more \textsc{Example} questions ($43.8\%$ of all samples). Yahoo dataset is more balanced, with \textsc{Procedural} questions being the most frequent type ($19.9\%$ of all samples). Distributions of question types for the two datasets are listed in Table~\ref{tab:data_question_type_dist} in Appendix~\ref{appendix:type_classifier}.

\section{Type-aware Open-ended Question Generation}
\label{sec:model}

\begin{figure}[t]
    \centering
    \includegraphics[width=0.47\textwidth]{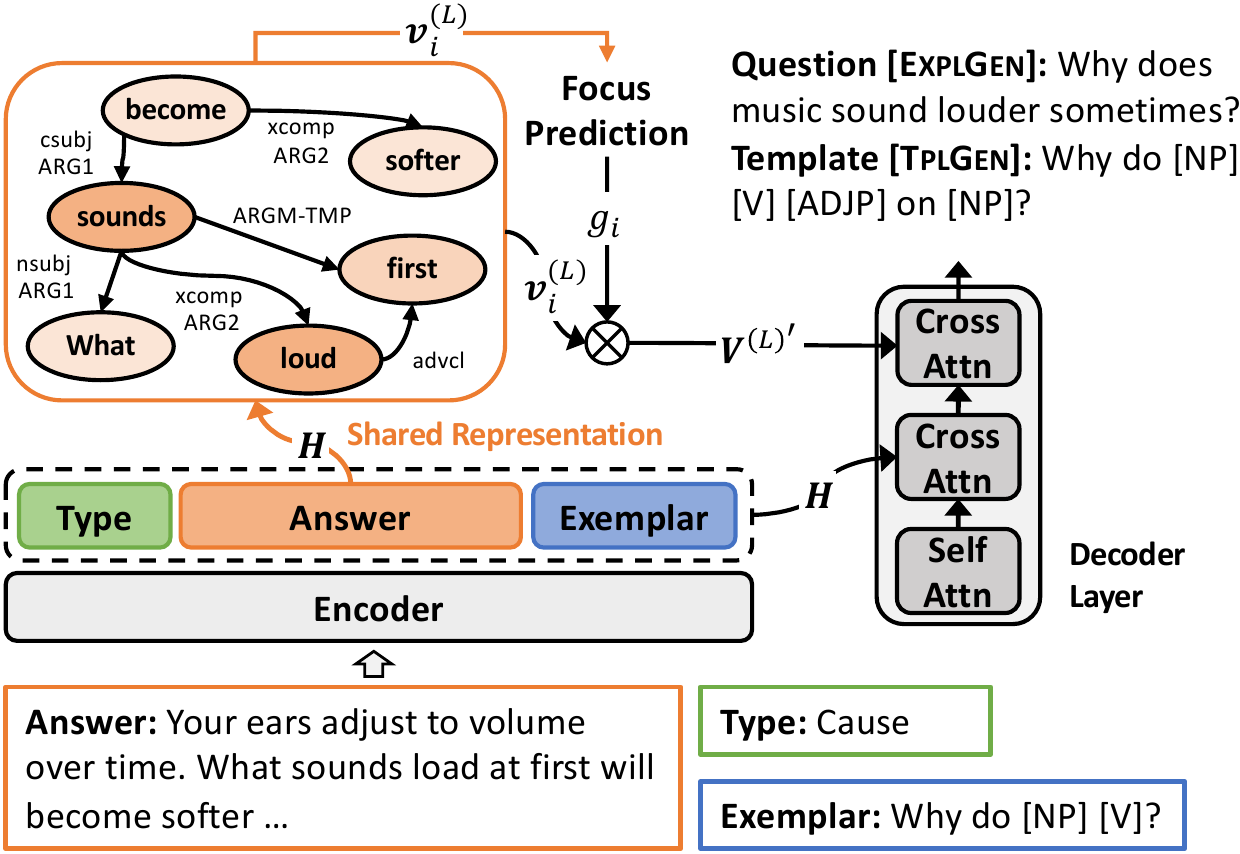}
    \caption{
    Our type-aware open-ended question generation framework. Detecting question focuses (nodes in darker color) and generating questions (or templates) are jointly learned. 
    We only show a partial semantic graph. Special tokens are also inserted to segment different parts of the input. 
    \textsc{JointGen} uses the type and the answer for question generation; 
    \textsc{ExplGen} further considers an exemplar; 
    and \textsc{TplGen} uses all three for template generation.}
    \label{fig:model}
\end{figure}

In this section, we present our type-aware question generation framework. As shown in Figure~\ref{fig:model}, our model takes in a multi-sentence text and a predicted question type. Built on shared input representations, it first detects question focuses from a semantic graph, and then generates the question (\S~\ref{subsec:method_joint}). We also propose two model variants that consider automatically extracted template exemplars or generated templates to achieve controllability (\S~\ref{subsec:template_gen}), enabling the generation of diverse questions.

\subsection{Joint Focus Prediction and Question Generation (\textsc{JointGen})}
\label{subsec:method_joint}

Our generator is built on top of BART~\cite{lewis-etal-2020-bart}. To facilitate the detection of salient content (i.e., focuses) to raise questions, we first augment the encoder with a semantic graph that consists of both dependency relations and semantic roles, capturing semantic relations over different scopes with varying granularities. Question focuses are first detected based on the semantic graph, which then guide question generation via cross-attentions, as shown in Figure~\ref{fig:model}. Although the joint modeling of focus prediction and question generation has been studied before, our design differs by using shared representations consisting of the input text and semantic graph, and the prediction of focuses are included through gating mechanisms, whereas previous work, e.g.~\newcite{pan-etal-2020-semantic}, simply employs multi-task learning. 
Below, we first describe constructing the semantic graph-augmented encoder, followed by the joint modeling of two tasks. 

\smallskip
\noindent \textbf{Improving Long Text Comprehension with Semantic Graph.} 
To construct the semantic graph, for each sentence, we start with obtaining its dependency tree using Stanford CoreNLP~\cite{manning-etal-2014-stanford}. To better highlight core concepts, we discard less important relations, e.g., auxiliaries. The full list is included in Appendix~\ref{appendix:graph_construction}. Since our goal is to detect central concepts that are well connected with many other words, we can remove relations on the edges to minimize the number of parameters to learn. 
Moreover, as semantic roles can indicate main entities~\cite{mannem2010question}, we extract semantic roles and their relations with AllenNLP~\cite{shi2019simple}. To merge the two sources of information, we add an edge in the dependency tree to connect the head word of the predicate and the head word of each semantic role.
To build a connected graph from the multi-sentence input, we add an edge between each sentence's last token and the next sentence's first token. 
Finally, we merge nodes with the same surface forms or with corefered mentions. 
To the best of our knowledge, this is the first time that both dependency and semantic relations are encoded in the same graph for question generation, and with enhanced connectivity of the constructed graph, our design can better signal content salience.

\smallskip 
\noindent \textbf{Joint Modeling with Cross-attentions.} 
Given a predicted question type $t$ and a multi-sentence text $x = \{ x_1, \cdots, x_n \}$, the BART encoder builds the contextual representation $\bm{H} = \{\bm{h}_0, \bm{h}_1, \cdots, \bm{h}_n \}$ at the last layer, where $\bm{h}_0$ is for $t$.

To encode the semantic graph, we initialize the node representation for node $v_i$ by taking the average contextual representations of its tokens and appending four bits encoding the number of nodes (capped at $10$) that are merged into $v_i$, to add frequency information. This step yields new node representations $\bm{v}^{(0)}_i$. We then apply graph attention networks (GATs)~\cite{velickovic2018graph} of $L$ layers to update the representations as follows:

\vspace{-1mm}
{
    \fontsize{10}{11}\selectfont
    \begin{align}
        \bm{v}^{(l)}_i &= \sum_{j \in \mathcal{N}_i} a_{i,j} \bm{W}^{(l)} \bm{v}^{(l-1)}_j
    \end{align}
}%
where $\bm{W}^{(l)}$ is a learnable parameter for the $l$-th layer, and $\mathcal{N}_i$ denotes the neighbors of $v_i$. The attention score $a_{i,j}$ is calculated as in GATs. We use $L=2$ for experiments. 

To \textbf{predict focuses}, the final node representation $\bm{v}^{(L)}_i$ is fed into the following feedforward network, yielding the probability of $v_i$ being a focus as: 

{
    \vspace{-2mm}
    \fontsize{10}{11}\selectfont
    \begin{align}
        p_{focus}(v_i = 1) = \sigma ( \bm{W}_1 \tanh (\bm{W}_2 \bm{v}^{(L)}_i) )
    \end{align}
}%
where $\bm{W}_1$ and $\bm{W}_2$ are learnable parameters. Bias terms are omitted for simplicity. 
We construct ground-truth labels by treating a node as a focus if it contains words used in the question.

To \textbf{generate the question}, we use the gating mechanism to inform the focus prediction results, where new node representations after being weighted by the focus probability are: 

{
    \fontsize{10}{11}\selectfont
    \noindent
    \begin{equation}
        \bm{v}^{(L)'}_i = g_i \odot \bm{v}^{(L)}_i \quad \quad g_i = p_{focus} (v_i = 1)
    \end{equation}
}%

Our model benefits from both large pre-training and hybrid semantic graphs by adding a separate cross attention for node presentations in each BART decoder layer. 
We then design \textbf{separate cross attentions} to attend (1) the output of the BART encoder, yielding $\bm{z}_e$, and (2) the node representations $\bm{V}^{(L)'}$, producing $\bm{z}_v$, which are formulated as: 

{
    \fontsize{10}{11}\selectfont
    \begin{align}
        \bm{z}_e &= \text{LN} (\bm{z}_s + \text{Attn}(\bm{z}_s, \bm{H})) \\
        \bm{z}_v &= \text{LN} (\bm{z}_e + \text{Attn}(\bm{z}_e, \bm{V}^{(L)'})) \\
        \bm{z'} &= \text{LN} (\bm{z}_v + \text{FFN}(\bm{z}_v))
    \end{align}
}%
where $\bm{z}_s$ denotes the output of self attentions for the current layer, and $\bm{z}'$ is the output for the layer. 
$\text{Attn}(\cdot, \cdot)$ denotes the multi-head attention operation as in \newcite{vaswani2017attention},  $\text{FFN}(\cdot)$ a feed-forward layer, and $\text{LN}(\cdot)$ is layer normalization.

Our final \textbf{training objective} accounts for both focus prediction and question generation objectives with equal weights.

\subsection{Diversifying Questions with Templates (\textsc{ExplGen} \& \textsc{TplGen})}
\label{subsec:template_gen}

An important goal of this work is to enable the generation of questions of diverse types. However, simply adding question type as input is insufficient (discussed in \S~\ref{sec:results}). We thus propose to leverage question templates to gain stronger controllability. Below we first present how to automatically extract templates from the training set, and then introduce two model variants that are built on the \textsc{JointGen} framework: \textsc{ExplGen} uses exemplar templates to guide the model to generate questions of selected types, and \textsc{TplGen} adds an extra step to first generate type-specific templates. 

\smallskip 
\noindent \textbf{Template Extraction.} 
While collecting templates specific to a given type, we need to ensure they remain topic-independent to be generalizable to different domains. 
To this end, we replace a word in the question with a template token that indicates its syntax function, e.g., \texttt{[V]} for a verb, if it appears in the answer after lemmatization. 
We further consider topically related words in the questions, by calculating word-level semantic similarities based on Numberbatch word embeddings~\cite{10.5555/3298023.3298212}, which are found to perform better on our datasets than other embeddings. 
Concretely, for each word in the answer, we replace the most similar word in the question with the template token.
This process is repeated until $80\%$ of content words in questions are replaced. 
Finally, for each noun phrase, adjective phrase, and adverb phrase, if its head word has been replaced, the whole phrase is transformed into a phrase type token. For instance, a question ``What are the differences between global warming and climate change?" becomes ``What are the differences between \texttt{[NP]} and \texttt{[NP]}?" 

\smallskip 
\noindent \textbf{Exemplars for Guidance (\textsc{ExplGen}).} 
Our first model variant considers adding a template exemplar for the given type as additional input, which provide more specific information to control the type of generated questions. Figure~\ref{fig:model} shows one such example. 
To identify exemplars, we use templates with frequencies above $20$ on Yahoo and $50$ on Reddit. We then manually inspect these templates and remove the ones with topic-specific words, resulting in $66$ exemplars for all types. They are listed in Table~\ref{tab:exemplar_list} in Appendix~\ref{appendix:templates_exemplars}.

During training, we choose the exemplar that has the lowest edit distance with the question, which is also used for training an exemplar selector based on RoBERTa. During testing, the exemplar with the highest selector score is used. The accuracy of the exemplar selector for each question type on the test set is reported in Table~\ref{tab:exemplar_classifier_acc} in Appendix~\ref{appendix:templates_exemplars}.

\smallskip 
\noindent \textbf{Generated Templates for Guidance (\textsc{TplGen}).} 
We further propose another model variant where we generate a new template and feed it (instead of an exemplar template as in \textsc{ExplGen}) as part of the question generation input. Specifically, we reuse \textsc{ExplGen} to learn to generate a target template, as derived from the template extraction procedure. 
During question realization, \textsc{TplGen} uses a BART-based generator that takes as input the question type, the input text, the generated template, and the words that are predicted as focuses. We use separate cross attentions to attend the representations of the focused words, similar to how node representations are attended in \textsc{JointGen}.

We recognize that having separate stages of exemplar selection and template generation introduces extra model training cost and potential errors in the pipeline. This work, however, focuses on improving the controllability as well as diversity of question generation, and we will leave the building of more efficient models in the future work.

\section{Experiment Results}
\label{sec:results}

\subsection{Automatic Evaluation}
\noindent \textbf{Comparisons and Metrics.} 
We compare with \textsc{DeepQG}~\cite{pan-etal-2020-semantic}, a model that uses dependency graphs for multi-hop question generation. We also compare with BART models that are fine-tuned on the same datasets as in our models, by using inputs of (1) the answer (\textsc{BART}), (2) the answer and a predicted question word (\textsc{BART+QWord}), and (3) the answer and a predicted question type (\textsc{BART+QType}). 
For \textsc{BART+QWord}, the question word is predicted by a RoBERTa classifier that considers the answer and is trained on our training sets. We follow \citet{liu2020asking} and use $9$ categories of question words. 
For both our models and \textsc{BART+QType}, the most confident type predicted by the classifier $\gamma_a$ (described in \S~\ref{subsec:question_type_ontology}), which reads in the answer, is used as input. 
To test the efficacy of semantic graphs, we further compare with a variant of \textsc{JointGen} that only uses the flat Transformer for focus prediction and question generation, denoted as \textsc{JointGen} w/o graph. 

We evaluate the generated questions with BLEU~\cite{papineni-etal-2002-bleu}, METEOR~\cite{lavie-agarwal-2007-meteor}, and ROUGE-L~\cite{lin-2004-rouge}.\footnote{We do not consider using Q-BLEU~\cite{nema-khapra-2018-towards} since it weighs question words highly.}

\begin{table}[t]
    
    \small
    \setlength{\tabcolsep}{0.4mm}
    \begin{tabular}{lllllll}
    \toprule
        & \multicolumn{3}{c}{\textit{Yahoo}} & \multicolumn{3}{c}{\textit{Reddit}} \\
        \textbf{Model} & \textbf{B-4} & \textbf{MTR} & \multicolumn{1}{c}{\textbf{R-L}} & \textbf{B-4} & \textbf{MTR} & \multicolumn{1}{c}{\textbf{R-L}}\\
        \midrule
        \textsc{DeepQG} & \hphantom{0}6.53 & 25.92 & 27.56 & \multicolumn{1}{c}{--} & \multicolumn{1}{c}{--} & \multicolumn{1}{c}{--} \\
        \textsc{BART} & 21.88 & 38.01 & 39.16 & 19.45 & 35.46 & 37.82 \\
        \textsc{BART+QWord} & 22.02 & 38.44 & 39.32 & 19.80$^\ast$ & \textbf{35.85} & 38.48$^\ast$ \\
        \hdashline
        \multicolumn{3}{l}{\textbf{Type-aware Models}} \\
        \textsc{BART+QType} & 22.12 & 38.62 & 39.72 & 19.90$^\ast$ & 35.83 & 38.68$^\ast$ \\

        \textsc{JointGen} (ours) & \textbf{22.56}$^\ast$ & \textbf{38.63} & \textbf{40.40}$^\ast$ & \textbf{20.09}$^\ast$ & 35.75 & \textbf{39.07}$^\ast$ \\
        \quad \quad \quad w/o graph & 22.21 & 38.21 & 39.93 & 19.81$^\ast$ & 35.60 & 38.47$^\ast$ \\
        \textsc{ExplGen} (ours) & 21.74 & 37.52 & 39.70 & 18.67 & 33.28 & 36.74 \\
        \textsc{TplGen} (ours) & 21.51 & 36.55 & 39.63 & 17.83 & 31.69 & 36.05 \\
        \bottomrule
    \end{tabular}
    \caption{
    Automatic evaluation results on Yahoo and Reddit with BLEU-4 (B-4), METEOR (MTR) and ROUGE-L (R-L). \textsc{JointGen} outperforms comparisons over all metrics except for METEOR on Reddit, but removing its graph degrades performance. 
    We only report results by \textsc{DeepQG} on Yahoo due to memory limitation. 
    $^\ast$: significantly better than BART ($p < 0.005$ with approximate randomization test).
    }
    \label{tab:qg_auto_result}
\end{table}

\smallskip
\noindent \textbf{Results} on both Yahoo and Reddit datasets are reported in Table~\ref{tab:qg_auto_result}. 
\textit{Our \textsc{JointGen} outperforms all comparisons on both datasets} over all automatic evaluation metrics except for METEOR on Reddit. 
When taking out the semantic graphs, model performance degrades substantially, which suggests that having structured representation is useful for focus detection and the final question generation task. 
We also observe a huge performance gap between \textsc{DeepQG} and systems based on BART, signifying the importance of leveraging pre-trained models for open-ended question generation. 
Meanwhile, \textit{adding question types helps BART generate more relevant questions} than using question words, indicating the value of our new question type ontology. 

Notably, our template-based generators, \textsc{ExplGen} and \textsc{TplGen}, which are trained to comply with the given templates, still produce comparable scores. This highlights \textit{the possibility to control the generated questions' types and syntax as demonstrated by the templates, without performance loss}.

\begin{table}[t]
    \small
    \centering
    \setlength{\tabcolsep}{0.5mm}
    \begin{tabular}{lllllll}
    \toprule
         & \multicolumn{3}{c}{\textit{Yahoo}} & \multicolumn{3}{c}{\textit{Reddit}} \\
        \textbf{Model} & \textbf{Acc}$\uparrow$ & \textbf{UnT}$\uparrow$ & \textbf{Pair}$\downarrow$ & \textbf{Acc}$\uparrow$ & \textbf{UnT}$\uparrow$ & \textbf{Pair}$\downarrow$ \\
        \midrule
        \textsc{BART} & \multicolumn{1}{c}{--} & 2.49 & \textbf{23.94} & \multicolumn{1}{c}{--} & 2.03 & \textbf{21.11} \\
        \textsc{BART+QWord} & \multicolumn{1}{c}{--} & 3.21 & 41.77 & \multicolumn{1}{c}{--} & 3.40 & 26.61 \\
        \hdashline
        \multicolumn{3}{l}{\textbf{Type-aware Models}} \\
        \textsc{BART+QType} & 22.47 & 2.27 & 44.39 & 23.67 & 2.20 & 80.91 \\
        \textsc{JointGen} & 43.90  & 4.03 & 63.06 & 22.75  & 2.33  & 60.72 \\
        \textsc{EXPLGen} & 75.47$^\ast$ & 6.98$^\ast$ & 25.12 & 65.79$^\ast$ & 6.23$^\ast$ & 22.17 \\ 
        \textsc{TPLGen} & \textbf{76.35}$^\ast$ & \textbf{7.08}$^\ast$ & 24.93 & \textbf{66.49}$^\ast$ & \textbf{6.32}$^\ast$ & 22.13 \\
         \bottomrule
    \end{tabular}
    \caption{
    Automatic evaluation on controllability and diversity by specifying $9$ different question types. We report type accuracy (Acc), number of unique types (UnT), and pairwise BLEU-4 (Pair).
    Our \textsc{ExplGen} and \textsc{TplGen} achieve stronger controllability by respecting the given question types more, as well as show higher diversity than comparisons except for BART with nucleus sampling. 
    $^\ast$: significantly better than all comparisons ($p < 0.005$).}
    \label{tab:control_qg_auto_result}
\end{table}

\smallskip
\noindent \textbf{Question Diversity Evaluation.} 
Next, we examine the controllability of models by specifying different question types as input. 
The top $9$ confident types\footnote{$9$ types are chosen because we only have $9$ categories of question words for \textsc{BART+QWord}.} predicted by our type predictor $\gamma_a$ are used as input to our models, producing $9$ questions for evaluation. For BART, we use nucleus sampling~\cite{Holtzman2020The} with $k = 10$ and $p = 0.7$ to sample diverse questions.

To evaluate, we first calculate the \textbf{question type accuracy} by comparing whether the types of the generated questions match the specified ones, with types labeled by our classifier $\gamma_q$ (\S~\ref{subsec:question_type_ontology}).
We then report the average numbers of \textbf{unique question types} in the $9$ generated questions per sample, with higher number indicating better controllability. 
Finally, we consider \textbf{pairwise BLEU-4}~\cite{cho-etal-2019-mixture} by computing the BLEU-4 between pairwise generated questions per sample, where lower values suggest higher content diversity. 

\begin{figure}[t]
    \centering
    \includegraphics[width=0.48\textwidth]{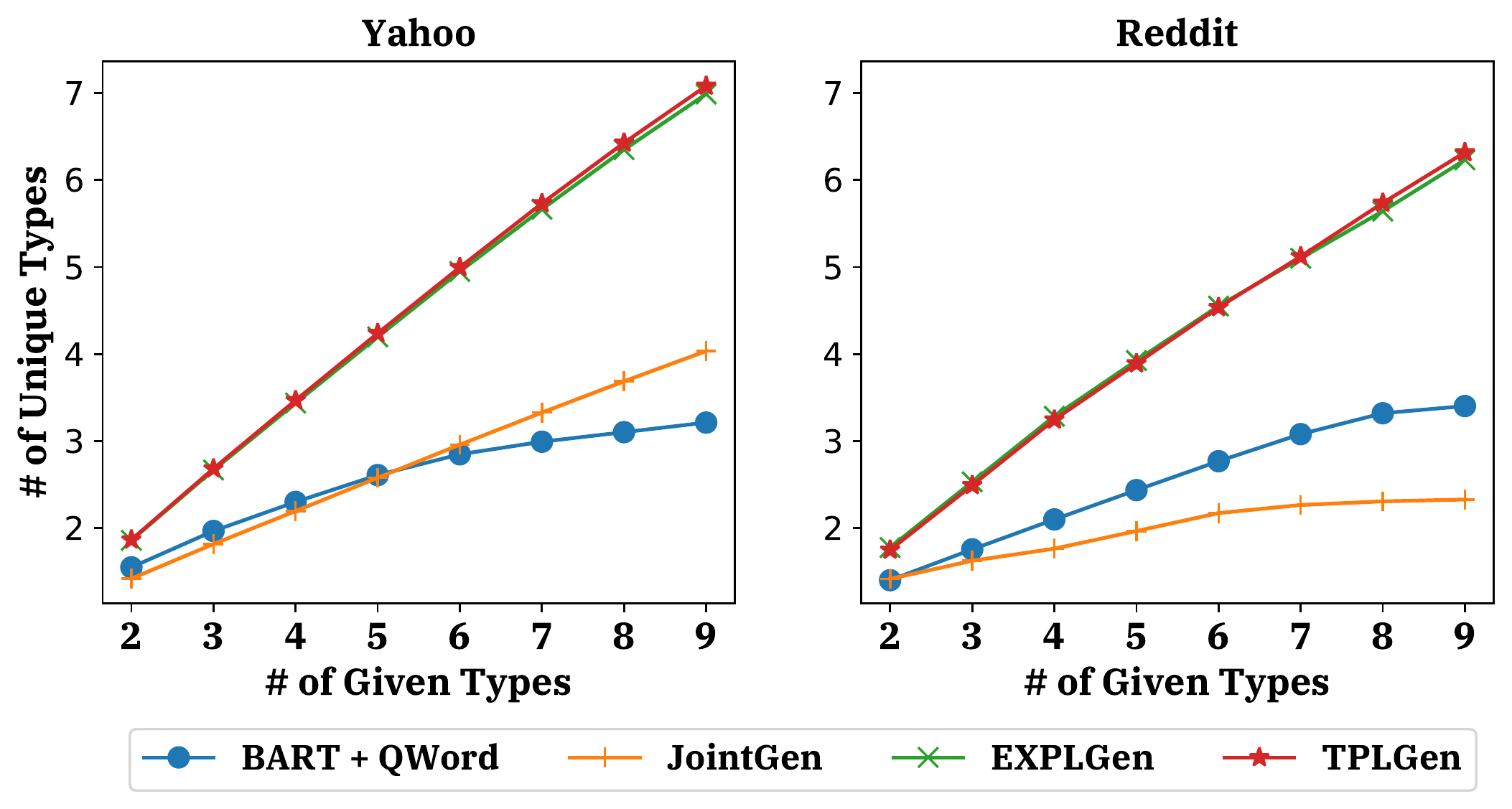}
    \vspace{-7mm}
    \caption{
    Number of unique types of the generated questions (Y-axis), when different numbers of question types are specified (X-axis).}
    \label{fig:compare_div_trend}
    \vspace{-2mm}
\end{figure}

First, \textit{our \textsc{ExplGen} and \textsc{TplGen} can generate questions with diverse types and content}, as shown by the significantly higher numbers of unique types than all comparisons and lower pairwise BLEU scores than comparisons except for BART with nucleus sampling in Table~\ref{tab:control_qg_auto_result}. This implies stronger type control by template-based generators, compared to \textsc{BART+QType} and \textsc{JointGen} which only use the question type token as input. Results on numbers of unique types by varying numbers of question types specified in the input are displayed in Figure~\ref{fig:compare_div_trend}, where \textsc{ExplGen} and \textsc{TplGen} maintain steady controllability.

Second, \textit{our question type ontology provides a new perspective for question diversity evaluation.} 
Among the comparisons, although BART with nucleus sampling and \textsc{BART+QWord} both have low pairwise BLEU, the types of questions they can generate are limited.

\subsection{Human Evaluation}

\noindent\textbf{Question Diversity.}
We hire three annotators who have participated in our question type annotation study to evaluate $80$ groups of questions generated by four selected models on each dataset.
For each group, we randomly sample an answer and indicate three most probably question types to each model, to generate three corresponding questions. 

\begin{table}[t]
    \begin{subtable}[h]{0.48\textwidth}
    \small
    \setlength{\tabcolsep}{0.7mm}
    \centering
    
    \begin{tabular}{lccc}
    \toprule
        \textbf{Model} & \textbf{Type} ($\%$) & \textbf{Syntax} ($\%$) & \textbf{Content} ($\%$) \\
        \midrule
        \textsc{BART} & 23.3 & 11.7 & 36.7 \\
        \textsc{BART+QWord} & 38.3 & 57.5 & 36.7 \\
        \textsc{ExplGen} & 77.9 & \textbf{61.2} & 58.3 \\
        \textsc{TplGen} & \textbf{81.2} & 58.3 & \textbf{64.2} \\
        \bottomrule
    \end{tabular}
    \caption{Yahoo}
    \end{subtable}
    
    \begin{subtable}[h]{0.48\textwidth}
    \small
    \setlength{\tabcolsep}{0.7mm}
    \centering
    
    \begin{tabular}{lccc}
    \toprule
        \textbf{Model} & \textbf{Type} ($\%$) & \textbf{Syntax} ($\%$) & \textbf{Content} ($\%$) \\
        \midrule
        \textsc{BART+QWord} & 40.0 & \textbf{50.0} & 22.9 \\
        \textsc{JointGen} & 63.3 & 31.2 & 20.4 \\
        \textsc{ExplGen} & \textbf{84.2} & 35.0 & 45.8 \\
        \textsc{TplGen} & 78.7 & 37.9 & \textbf{60.4} \\
        \bottomrule
    \end{tabular}
    \caption{Reddit}
    \end{subtable}
    \vspace{-2mm}
    \caption{
    Percentage of samples marked as having the most diverse question types, syntax structures, and content of answers. Ties are allowed. The Krippendorf's $\alpha$s for the three aspects are $0.56$, $0.43$, and $0.25$ on Yahoo, and are $0.54$, $0.38$, and $0.38$ on Reddit.}
    \label{tab:control_qg_human_result}
    \vspace{-2mm}
\end{table}

For each sample, the annotators are asked to rank the four models from 1 (highest) to 4 (lowest) on three aspects of diversities: \textbf{type}--whether the three generated questions have different types, \textbf{syntax}--whether they use different syntax, and \textbf{answer content}--whether the three questions need to be addressed with different answers. Ties are allowed.

We find that \textit{human judges rate questions generated by our \textsc{ExplGen} and \textsc{TplGen} as having greater diversities over all aspects}, except for syntax diversity on Reddit, as shown in Table~\ref{tab:control_qg_human_result}. 
Among the two model variants, questions by \textsc{TplGen} yield more diverse answers. Based on our observation, \textsc{TplGen} uses automatically generated templates to produce more focused questions with different answers, compared to \textsc{ExplGen} which employs exemplars. This shows the promise of using automatically generated templates to create questions that need to be addressed with different answers. Besides Figure~\ref{fig:intro}, we show more sample outputs in Figure~\ref{fig:sample_output}, where \textsc{ExplGen} and \textsc{TplGen} exhibit stronger controllability than \textsc{JointGen}.

\begin{figure}[t]
    \centering
    \fontsize{9}{11}\selectfont
    \setlength{\tabcolsep}{0.8mm}
    \begin{tabular}{p{0.46\textwidth}}
    \toprule
        \textbf{Answer:} My sister in law and her husband ``genetically modified'' their second child because the first has EB. They eliminated that and had a baby that gets to live pain free. Under the right circumstances, I'm all for it ...  \\
        \hline
        \textbf{\textsc{JointGen}} \\
        \st{\textsc{[Procedural]}} How would you feel about genetically modified babies? \\
        \textsc{[Judgmental]} What are your thoughts on genetically modified babies? \\
        \textsc{[Verification]} Is it possible to genetically modify a child? \\
        \hline
        \textbf{\textsc{ExplGen}} \\
        \textsc{[Procedural]} \hlc[cyan!20]{How do} people genetically modify their children? \\
        \textsc{[Judgmental]}\hlc[cyan!20]{What do you think} about genetically modified babies? \\
        \textsc{[Verification]} \hlc[cyan!20]{Can} you genetically modify a baby to have a better chance of survival? \\
        \hline
        \textbf{\textsc{TplGen}} \\
        \textsc{[Procedural]} \hlc[cyan!20]{How do people} genetically modify their children? \\
        \textsc{[Judgmental]} \hlc[cyan!20]{What do you think about} genetically modified babies? \\
        \textsc{[Verification]} \hlc[cyan!20]{Can you} genetically modify a baby? \\
        \bottomrule
    \end{tabular}
    \caption{
    Sample outputs of our models given different question types. 
    Spans that belong to the exemplars or the generated templates are colored with \hlc[cyan!20]{blue}. 
    Generated questions that do not match the given type are marked by \st{strikethrough}.}
    \label{fig:sample_output}
    \vspace{-2mm}
\end{figure}

\smallskip
\noindent\textbf{Question Content Quality.}
We use the same set of human judges to evaluate another $80$ groups of questions output by five selected models and the reference. Three aspects are rated from 1 (worst) to 5 (best): \textbf{appropriateness}--whether the question is semantically correct, \textit{without} considering the answer; \textbf{answerability}--whether the question can be addressed by the given answer; and \textbf{scope}--whether the question is related to a longer span of the answer (global scope) or focuses on local content (e.g., one phrase or one sentence). 
We further ask the annotators to rank questions based on their overall quality and preferences, with ties allowed.

As shown in Table~\ref{tab:qg_human_result}, our \textsc{JointGen} model produces questions with better answerability and that cover broader content in the answers. It is also rated as the best in more than half of the evaluation instances on both datasets. 
Between \textsc{BART+QWord} and \textsc{BART+QType}, human judges rate the system outputs that conditioned on our question types to have better overall quality.

\begin{table}[t]
    \begin{subtable}[h]{0.48\textwidth}
    \centering
    \small
    \begin{tabular}{lccccc}
    \toprule
        \textbf{Model} & \textbf{Appro.} & \textbf{Ans.} & \textbf{Scp.} & \textbf{Top 1} \\
        \midrule
        \textsc{Reference} & 4.77 & 3.96 & 3.79 & 34.5$\%$ \\
        \midrule
        \textsc{BART} & \textbf{4.93} & 4.02 & 3.81 & 39.7$\%$ \\
        \textsc{BART+QWord} & 4.86 & 4.14 & 3.85 & 40.8$\%$ \\
        \textsc{BART+QType} & 4.92 & 4.23 & 3.94 & 48.7$\%$ \\
        \textsc{JointGen} & 4.90 & \textbf{4.25} & \textbf{3.96} & \textbf{50.5}$\%$ \\
        \textsc{TPLGen} & 4.92 & 4.19 & 3.87 & 46.4$\%$ \\
        \bottomrule
    \end{tabular}
    \caption{\textsc{Yahoo}}
    \end{subtable}
    
    \begin{subtable}[h]{0.48\textwidth}
    \centering
    \small
    \begin{tabular}{lccccc}
    \toprule
        \textbf{Model} & \textbf{Appro.} & \textbf{Ans.} & \textbf{Scp.} & \textbf{Top 1} \\
        \midrule
        \textsc{Reference} & 4.90 & 4.43 & 4.37 & 47.1$\%$ \\
        \midrule
        \textsc{BART} & \textbf{4.89} & 4.27 & 4.21 & 43.9$\%$ \\
        \textsc{BART+QWord} & 4.88 & 4.29 & 4.21 & 46.7$\%$ \\
        \textsc{BART+QType} & 4.88 & 4.39 & 4.26 & 49.6$\%$ \\
        \textsc{JointGen} & 4.84 & \textbf{4.45} & \textbf{4.38} & \textbf{50.3}$\%$ \\
        \textsc{TPLGen} & 4.81 & 4.21 & 4.19 & 33.1$\%$ \\
        \bottomrule
    \end{tabular}
    \caption{\textsc{Reddit}}
    \end{subtable}
    \caption{Human evaluation on appropriateness (Appro.), answerability (Ans.), scope of answers (Scp.), and percentage of questions rated as having the best overall quality (Top 1), with ties allowed.  \textsc{JointGen} is rated as with higher answerability and with answers requiring a broader scope. 
    The Krippendorff's $\alpha$s for the four aspects are $0.42$, $0.56$, $0.50$, $0.39$ on Yahoo, and are $0.34$, $0.46$, $0.43$, $0.35$ on Reddit.}
    \label{tab:qg_human_result}
    \vspace{-2mm}
\end{table}

\subsection{Further Analyses}

\begin{figure}[t]
    \centering
    \begin{subfigure}[b]{0.48\textwidth}
        \centering
        \includegraphics[width=\textwidth]{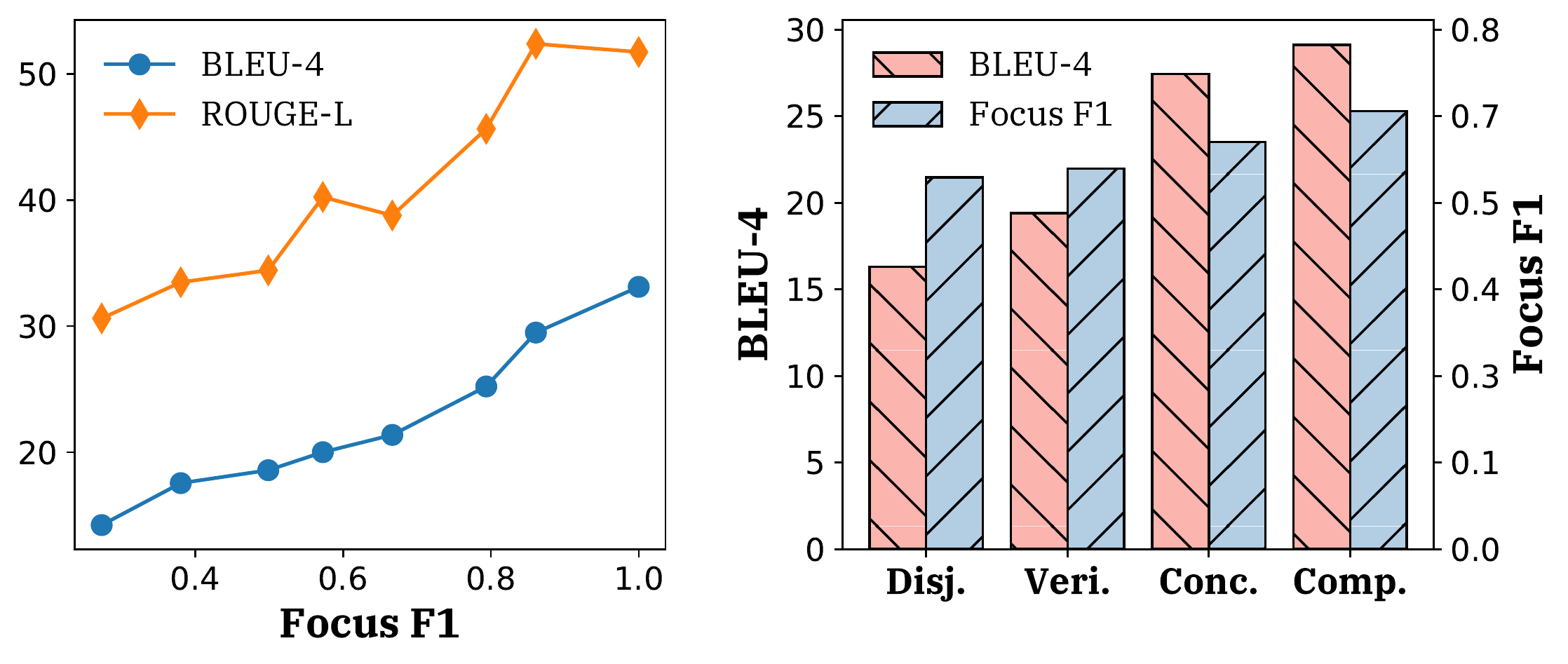}
        \caption{Yahoo}
    \end{subfigure}
    \begin{subfigure}[b]{0.48\textwidth}
        \centering
        \includegraphics[width=\textwidth]{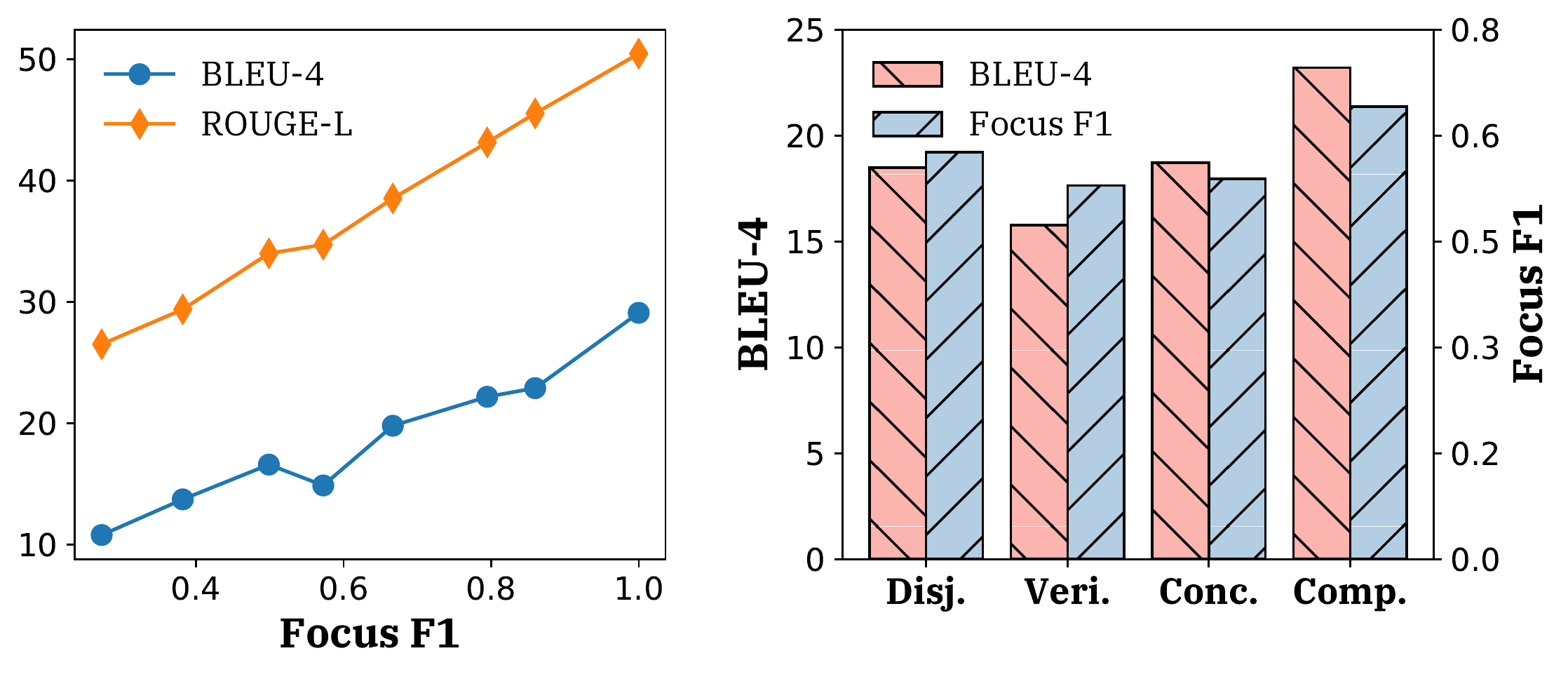}
        \caption{Reddit}
    \end{subfigure}
    \vspace{-8mm}
    \caption{ 
    On both Yahoo and Reddit, we find positive effect of focus prediction on question quality (measured by BLEU-4 and ROUGE-L) (left), which is also broken down by question types (right). Disj.: \textsc{Disjunctive}, Veri.: \textsc{Verification}, Conc.: \textsc{Concept}, Comp.: \textsc{Comparison}. 
    }
    \label{fig:qt_breakdown_perf}
    \vspace{-3mm}
    
\end{figure}

\noindent\textbf{Does focus prediction correlate with question quality?} 
We first investigate the relationship between focus prediction and question generation by using our joint model \textsc{JointGen}.
As can be seen from Figure~\ref{fig:qt_breakdown_perf}, there is a strong correlation between F1 scores of focus prediction and BLEU-4 as well as ROUGE-L, where samples in the Yahoo and Reddit test sets are grouped into 8 bins based on the F1 scores. 
The Pearson correlation coefficients between BLEU-4 and focus F1 are $0.29$ on Yahoo and $0.26$ on Reddit. For ROUGE-L, the correlation coefficients are $0.35$ on Yahoo and $0.34$ on Reddit. All the correlations have $p < 10^{-5}$. The strong positive correlations imply the importance of accurate focus prediction for open-ended question generation. We also show the F1 scores and BLEU-4 for selected question types on the right of Figure~\ref{fig:qt_breakdown_perf}, again demonstrating the effect of focus detection on question quality.

\smallskip
\noindent\textbf{When do our models fail to respect the given types?} 
Next, we provide insights into which types of questions are challenging to generate by using our template-based models \textsc{ExplGen} and \textsc{TplGen}. Both variants frequently fail to respect the given question type of \textsc{Verification}, in which cases they often produce \textsc{Judgemental} questions. They also tend to confuse \textsc{Example} and \textsc{Extent} with \textsc{Concept} questions. 
After manually inspecting $50$ generated questions for the aforementioned three types, we find that many of them can be labeled with both types, thus creating confusion for our classifier. For instance, \textit{``What are the import restrictions in the US?''} can be considered as either asking for a definition or for examples. Therefore, future work should include designing multi-class type identification models.

\section{Conclusion}

We present a new question type ontology which better captures the nuances of questions to support the study of open-ended question generation. We further annotate a new dataset with $4{,}959$ questions based on the proposed ontology. We describe a joint question focus detection and question generation framework with a novel semantic graph-augmented representation, which is directly built on large pre-trained models. Based on this framework, we also enhance the controllability and diversity of generated questions by employing template exemplars or automatically generated templates. Experiments on two large datasets show that questions generated by our models have better quality and higher diversity than non-trivial comparisons, with similar results rated by human judges.

\section*{Acknowledgements}
This research is supported in part by National Science Foundation through Grants IIS-1813341 and a CAREER award IIS-2046016. 
We thank three anonymous reviewers, area chair, and senior area chairs for their valuable suggestions for improving various aspects of this work. 
\section*{Ethics Statement}
Large models that are pre-trained on heterogeneous web data are shown to encode biases and can be potentially harmful for marginalized populations. 

While the automatically learned templates improve controllability in question generation, we also recognize that our system might be misused to create questions that contain objectionable content. We therefore advocate cautious and responsible practices in real-world deployment.

Our data collection process for the two new datasets involves removing samples with abusive languages and human inspection on random samples. Given the data volume, however, we cannot exhaustively verify that all records are free of potentially offensive content.

\bibliographystyle{acl_natbib}
\bibliography{acl2021}

\newpage
\appendix



\section{Data Collection}
\label{appendix:data_collection}

\paragraph{Data Filtering.}

After collecting the raw data from Yahoo and Reddit, we design rules to filter out ill-formed answers and questions. These rules are listed in Table~\ref{tab:data_clean_rule}. 
Finally, we conduct human inspection on random samples from the two datasets and confirm that samples are all clean and contain open-ended questions.

\begin{table}[ht]
    \centering
    \small
    \setlength\extrarowheight{1pt}
    \begin{tabular}{p{0.45\textwidth}}
    \toprule
        \textbf{Rules for Data Cleaning} \\
        \midrule
        - The question has URL links. \\
        - The question has more than 1 sentence or does not end with a question mark. \\
        - The question has less than 4 words or less than 1 content word. \\
        - The question does not start with wh-words: \textit{what, why, how, which, where, who, when}; yes-no words: \textit{is, are, was,
             were, will, would, do, does, did, can, could, should, has, have}; or frequent words for conditions: \textit{if, in, for, to, as, at}. \\
        \midrule
        - The answer has less than 15 content words. \\
        - The answer has less content words than the question. \\
        - The answer has more than 30\% of the words as digit letters. \\
        \midrule
        - The question and the answer have less than 2 overlapping content words. \\
        - The question or the answer contains abusive words from Google's ``what do you need'' project\tablefootnote{\url{https://gist.github.com/jamiew/1112488}}. \\
        - The question or the answer has emoticons\tablefootnote{\url{https://en.wikipedia.org/wiki/List\_of\_emoticons}}. \\
        - The question or the answer has 3 consecutive punctuation. \\
        - The question or the answer has 3 consecutive fully uppercased words. \\
        - The question has more than 90\% of title-case words or the answer has more than 30\% of title-case words. \\
        - The question has more than 1 unique word not in the English dictionary or the answer has more than 2 unique words not in the English dictionary\tablefootnote{\url{https://github.com/dwyl/english-words}}. \\
        \bottomrule
    \end{tabular}
    \caption{Rules for filtering out ill-formed question-answer pairs on both Yahoo and Reddit.}
    \label{tab:data_clean_rule}
\end{table}

\paragraph{Question Type Annotation.} 

We include the definition and corresponding examples for each question type in the annotation guideline, as shown in Table~\ref{fig:ques_type_anno_guideline}. 
We allow annotators to label a question with two types if they cannot decide between the two. All recruited annotators are U.S. college students, and are paid \$$15$ per hour for the task. On average, it takes $3.5$ hours to annotate $1000$ questions.

For samples with disagreed labels, we check whether agreement can be reached by considering both labeled types. For example, if annotator A labels \textsc{Verification} and \textsc{Judgmental}, and annotator B labels \textsc{Judgmental}, the agreed-upon type is \textsc{Judgmental}.
We then resolve outstanding disagreements by discussion.


\section{Details for Question Type Classifiers}
\label{appendix:type_classifier}

To train the question type classifier $\gamma_q$ that reads the question as input, we split the collected question type dataset into training, validation, and test sets. Sample counts and question type distributions for different data splits are shown in Table~\ref{tab:qt_data_split}.

\begin{table}[th]
    \centering
    \small
    \begin{tabular}{lccc}
    \toprule
        \textbf{Question Type} & \textbf{Training} & \textbf{Validation} & \textbf{Test} \\
        \midrule
        \textsc{Verification} & 445 & 58 & 72 \\
        \textsc{Disjunctive} & 156 & 36 & 36 \\
        \textsc{Concept} & 289 & 46 & 54 \\
        \textsc{Extent} & 274 & 49 & 48 \\
        \textsc{Example} & 871 & 152 & 139 \\
        \textsc{Comparison} & 162 & 25 & 30 \\
        \textsc{Cause} & 475 & 69 & 91 \\
        \textsc{consequence} & 102 & 14 & 11 \\ 
        \textsc{Procedural} & 469 & 63 & 85 \\
        \textsc{Judgmental} & 476 & 68 & 94 \\
        \midrule
        \textsc{All} & 3719 & 580 & 660 \\
        \bottomrule
    \end{tabular}
    \caption{
    Sample counts and question type distributions for the newly labeled dataset with question types.}
    \label{tab:qt_data_split}
\end{table}

We then use $\gamma_q$ to identify types for unlabeled questions in Yahoo and Reddit. The question type distributions for the two datasets are shown in Table~\ref{tab:data_question_type_dist}.

\begin{table}[th]
    \centering
    \small
    \begin{tabular}{lcccc}
    \toprule
        & \multicolumn{2}{c}{\textit{Yahoo}} & \multicolumn{2}{c}{\textit{Reddit}} \\
        \textbf{Question Type} & \textbf{\#} & \textbf{\%} & \textbf{\#} & \textbf{\%} \\
        \midrule
        \textsc{Verification} & 48{,}577 & 16.7 & 43{,}801 & 6.1 \\
        \textsc{Disjunctive} & 5{,}131 & 1.8 & 7{,}179 & 1.0 \\
        \textsc{Concept} & 44{,}963 & 15.5 & 34{,}744 & 4.8 \\
        \textsc{Extent} & 16{,}811 & 5.8 & 19{,}217 & 2.7 \\
        \textsc{Example} & 29{,}494 & 10.1 & 315{,}478 & 43.8 \\
        \textsc{Comparison} & 13{,}167 & 4.5 & 14{,}727 & 2.0 \\
        \textsc{Cause} & 35{,}010 & 12.0 & 128{,}204 & 17.8 \\
        \textsc{consequence} & 5{,}547 & 1.9 & 16{,}542 & 2.3 \\ 
        \textsc{Procedural} & 57{,}762 & 19.9 & 66{,}662 & 9.3 \\
        \textsc{Judgmental} & 34{,}149 & 11.8 & 73{,}434 & 10.2 \\
        \bottomrule
    \end{tabular}
    \caption{Distributions of question types for the two datasets as labeled by $\gamma_q$.}
    \label{tab:data_question_type_dist}
\end{table}


\section{Details for Graph Construction}
\label{appendix:graph_construction}

We discard secondary dependency relations for graph construction, including \textit{case, mark, cc, cc:preconj, aux, aux:pass, cop, det, discourse, expl, det:predet, punct, ref}. The definition for each dependency can be found in Universal Dependency.\footnote{\url{https://universaldependencies.org/}}


\section{Details for Templates and Exemplars}
\label{appendix:templates_exemplars}

\paragraph{Template Construction.} 
To avoid replacing words that are representative of question types during template construction, we maintain a list of words not to be replaced for each question type, as shown in Table~\ref{tab:non_replace_word}. These words are identified by frequency with additional manual inspection.

\begin{table}[th]
    \centering
    \small
    \setlength{\tabcolsep}{0.9mm}
    \begin{tabular}{lp{0.33\textwidth}}
    \toprule
        \textbf{Question Type} & \textbf{Words Not to Be Replaced} \\
        \midrule
        \textsc{Verification} & - \\
        \textsc{Disjunctive} & \textit{or} \\
        \textsc{Concept} & \textit{mean} \\
        \textsc{Extent} & \textit{many, much, long, take, get} \\
        \textsc{Example} & \textit{good, best, find, anyone, get} \\
        \textsc{Comparison} & \textit{difference, best, better, and, or} \\
        \textsc{Cause} & \textit{people} \\
        \textsc{consequence} & \textit{happen, happens, would, affect, effect, effects} \\ 
        \textsc{Procedural} & \textit{get, way, make, best, know} \\
        \textsc{Judgmental} & \textit{think, would, like, anyone} \\
        \bottomrule
    \end{tabular}
    \caption{Words not to be replaced during template construction, per question type.}
    \label{tab:non_replace_word}
\end{table}

\paragraph{Exemplar Collection.}

Table~\ref{tab:exemplar_list} lists the collected template exemplars for different question types. 

\begin{table}[t!]
    \centering
    \small
    \setlength\extrarowheight{6pt}
    \setlength{\tabcolsep}{0.9mm}
    \begin{tabular}{lp{0.33\textwidth}}
        \toprule
        \textbf{Question Type} & \textbf{Template Exemplars} \\
        \midrule
        \textsc{Verification} & ``Is \texttt{[NP]} \texttt{[NP]}?'', ``Is there \texttt{[NP]}?'', ``Is \texttt{[NP]} \texttt{[ADJP]}?'', ``Can \texttt{[NP]} \texttt{[V]} \texttt{[NP]}?'', ``Do \texttt{[NP]} \texttt{[V]} \texttt{[NP]}?'', ``Does anyone have \texttt{[NP]}?'', ``Is it \texttt{[ADJP]} to \texttt{[V]} \texttt{[NP]}?''\\
        \textsc{Disjunctive} & ``Is \texttt{[NP]} \texttt{[NP]} or \texttt{[NP]}?'', ``Is \texttt{[NP]} \texttt{[ADJP]} or \texttt{[ADJP]}?'', ``Who is \texttt{[NP]} or \texttt{[NP]}?'', ``What came \texttt{[ADVP]} \texttt{[NP]} or \texttt{[NP]}?'', ``Which is \texttt{[NP]} or \texttt{[NP]}?'', ``What is \texttt{[NP]} or \texttt{[NP]}?'' \\
        \textsc{Concept} & ``What is \texttt{[NP]}?'', ``What does \texttt{[NP]} mean?'', ``Who is \texttt{[NP]}?'', ``Where is \texttt{[NP]}?'', ``What is the meaning of \texttt{[NP]}?'', ``What does \texttt{[NP]} do?'', ``What do you know about \texttt{[NP]}?'', ``When is \texttt{[NP]}?'', ``What is meant by \texttt{[NP]}?'', ``Where did \texttt{[NP]} come from?'', ``Which is \texttt{[NP]}?'', ``When was \texttt{[NP]} \texttt{[V]}?'', ``What is the definition of \texttt{[NP]}?'', ``How is \texttt{[NP]}?'', ``Does anyone know anything about \texttt{[NP]}?'', ``What happened to \texttt{[NP]}?'' \\
        \textsc{Extent} & ``What is \texttt{[NP]}?'', ``How \texttt{[OTHER]} is \texttt{[NP]}?'', ``How many \texttt{[OTHER]} are in \texttt{[NP]}?'', ``How many \texttt{[NP]}?'', ``How much does \texttt{[NP]}?'' \\
        \textsc{Example} & ``What are \texttt{[NP]}?'', ``What is a good \texttt{[NP]}?'', ``What is the best \texttt{[NP]}?'', ``Where can I \texttt{[V]} \texttt{[NP]}?'', ``What are some good \texttt{[NP]}?'', ``Does anyone have \texttt{[NP]}?'' \\
        \textsc{Comparison} & ``What is the difference between \texttt{[NP]} and \texttt{[NP]}?'', ``What is the best \texttt{[NP]}?'', ``What is \texttt{[NP]}?'', ``Which is better \texttt{[NP]} or \texttt{[NP]}?'', ``Who is the best \texttt{[NP]}?'' \\
        \textsc{Cause} & ``Why is \texttt{[NP]}?'', ``Why is \texttt{[NP]} \texttt{[ADJP]}?'', ``Why do \texttt{[NP]}?'', ``Why do \texttt{[NP]} \texttt{[V]} \texttt{[NP]}?'', ``What causes \texttt{[NP]}?'', ``Why do \texttt{[NP]} \texttt{[V]}?'' \\
        \textsc{consequence} & ``What are \texttt{[NP]}?'', ``How does \texttt{[NP]} affect \texttt{[NP]}?'', ``What are \texttt{[NP]} effects \texttt{[NP]}?'', ``What are the benefits of \texttt{[NP]}?'' \\ 
        \textsc{Procedural} & ``How do I \texttt{[V]} \texttt{[NP]}?'', ``How to \texttt{[V]} \texttt{[NP]}?'', ``How do you \texttt{[V]} \texttt{[NP]}?'', ``What is the best way to \texttt{[V]} \texttt{[NP]}?'', ``How is \texttt{[NP]} \texttt{[V]}?'', ``How does \texttt{[NP]} \texttt{[V]} \texttt{[NP]}?'', ``How do \texttt{[NP]} work?'' \\
        \textsc{Judgmental} & ``What is \texttt{[NP]}?'', ``What do you think of \texttt{[NP]}?'', ``Do you believe in \texttt{[NP]}?'', ``Do you like \texttt{[NP]}?'', ``Should I \texttt{[V]} \texttt{[NP]}?'', ``Is \texttt{[NP]} \texttt{[NP]}?'', ``Do you \texttt{[V]} \texttt{[NP]}?'', ``Who is \texttt{[NP]}?'', ``Are you \texttt{[NP]}?'', ``Should \texttt{[NP]} \texttt{[V]} \texttt{[NP]}?'' \\
        \bottomrule
    \end{tabular}
    \caption{Template exemplars for different question types.}
    \label{tab:exemplar_list}
\end{table}

\paragraph{Exemplar Classifiers.}

To predict the exemplars used for question decoding, we train one exemplar classifier for each question type, on each dataset. Accuracy values of these exemplar classifiers on the reserved test sets are listed in Table~\ref{tab:exemplar_classifier_acc}.

\begin{table}[t]
    \centering
    \small
    \begin{tabular}{lcc}
    \toprule
        \textbf{Question Type} & \textbf{Yahoo Acc} & \textbf{Reddit Acc} \\
        \midrule
        \textsc{Verification} & 43.04 & 38.09 \\
        \textsc{Disjunctive} & 65.21 & 47.50 \\
        \textsc{Concept} & 43.43 & 43.84 \\
        \textsc{Extent} & 45.56 & 26.99 \\
        \textsc{Example} & 46.06 & 50.26 \\
        \textsc{Comparison} & 53.25 & 54.33 \\
        \textsc{Cause} & 54.58 & 43.77 \\
        \textsc{consequence} & 38.94 & 26.57 \\ 
        \textsc{Procedural} & 43.66 & 29.55 \\
        \textsc{Judgmental} & 28.31 & 28.24 \\
        \bottomrule
    \end{tabular}
    \caption{Accuracy of exemplar classifiers for different question types on Yahoo and Reddit.}
    \label{tab:exemplar_classifier_acc}
\end{table}

\section{Details for Implementation}

We use Fairseq~\cite{ott2019fairseq} to build our models and conduct training and decoding. For the Graph Attention Networks (GATs) in our focus predictor, we adopt the implementation by PyTorch Geometric~\cite{Fey/Lenssen/2019}. All our experiments are conducted on a Quadro RTX 8000 GPU with 48 GB of memory.

\paragraph{Training Settings.}

We use Adam~\cite{kingma2014adam} for the training of all our models. Our question type classifiers and template exemplar classifiers are trained with a maximum learning rate of $1 \times 10^{-5}$ and a batch size of $32$. For training generation models, the maximum learning rate is $3 \times 10^{-5}$ and each batch contains at most $32{,}768$ tokens. Mixed-precision training is adopted for all models except for models with GATs.

\paragraph{Decoding Settings.}

We use beam search for decoding. A beam size of $5$ and a length penalty of $1.5$ are used for all models. Repeated trigram blocking is applied to question generation. The minimum and maximum lengths for generation are set to $1$ and $100$, respectively.

\paragraph{Model Parameters.}

The question type classifiers and template exemplar classifiers are based on RoBERTa$_{\mathrm{Large}}$, which has $355$M parameters. Our generation model builds a GAT upon the BART model, containing $430$M parameters in total.

\paragraph{Running Time.}

Training question type classifiers takes $23$ hours. Due to the difference in training data size, the training time for template exemplar classifiers ranges from $20$ minutes to $3$ hours. For our generation model with focus prediction, it takes $6$ hours to train on Yahoo and $12$ hours to train on Reddit. Decoding on the test set of Yahoo and Reddit takes $8$ minutes and $15$ minutes, respectively.



\begin{table*}[t]
    \centering
    \small
    \setlength{\tabcolsep}{0.7mm}
    \begin{tabular}{|p{0.94\textwidth}|}
        \hline
        In this study, you are asked to annotate the question types for 1000 questions. The question type reflects the nature of the question. It is \textbf{not determined by the interrogative word} of the question. There are 10 question types in total. The definition for each type is shown in the following Table, along with examples per question type. \\
        During annotation, you can label two most-confident types when no clear decision can be made for the most probable type. 
        \\
        \hline
        \hline
        \textsc{\textbf{Verification}}:
        Asking for the truthfulness of an event or a concept. \\
        - \textit{``Is Michael Jackson an African American?''} \\
        - \textit{``Does a Mercedes dealer have to unlock a locked radio?''} \\
        - \textit{``Could stress, anxiety, or worry cause cholesterol levels to rise?''} \\
        
        \hline
        \textsc{\textbf{Disjunctive}}:
        Asking for the true one given multiple events or concepts, where comparison among options is not needed. \\
        - \textit{``Is Michael Jackson an African American or Latino?''} \\
        - \textit{``Is a DVI to HDMI cable supposed to transmit audio and video or just video?''} \\
        - \textit{``When you get a spray-on tan does someone put it on you or does a machine do it?''}\\
        
        \hline
        \textsc{\textbf{Concept}}:
        Asking for a definition of an event or a concept. \\
        - \textit{``Who said the sun never sets on the British empire?''} \\
        - \textit{``Where do dolphins have hair at?''} \\
        - \textit{``What is the origin of the phrase "kicking the bucket"?''} \\
        
        \hline
        \textsc{\textbf{Extent}}:
        Asking for the extent or quantity of an event or a concept. \\
        - \textit{``How long does gum stay in your system?''} \\
        - \textit{``What is Barry Larkin's hat size?''} \\
        - \textit{``To what extent is the Renewable Fuel Standard accurate nationwide?''} \\
        
        \hline
        \textsc{\textbf{Example}}:
        Asking for example(s) or instance(s) of an event or a concept. \\
        - \textit{``What are some examples to support or contradict this?''} \\
        - \textit{``Where can I get my teeth examined around Los Angeles?''} \\
        - \textit{``What countries/regions throughout the world do not celebrate the Christmas holidays?''} \\
        - \textit{``What is the best goal or win you have ever made in a sport?''} \\
        
        \hline
        \textsc{\textbf{Comparison}}:
        Asking for comparison among multiple events or concepts. \\
        - \textit{``How does an electric violin "play" differently than an acoustic violin?''} \\
        - \textit{``What is the best tinted facial moisturizer?''} \\
        - \textit{``In what hilariously inaccurate ways is your job/career portrayed on television or in movies?''} \\
        - \textit{``Which is better, Nike or Adidas?''} \\
        
        \hline
        \textsc{\textbf{Cause}}:
        Asking for the cause or reason for an event or a concept. \\
        - \textit{``How does the D.M.V. decide the first letter of the California driver’s license?''} \\
        - \textit{``Why are parents strick on girls than boys?''} \\
        - \textit{``What makes nerve agents like "Novichok" so hard to produce and why can only a handful of laboratories create them?''} \\
        \textit{``Why is the sky blue?''} \\
        
        \hline
        \textsc{\textbf{Consequence}}:
        Asking for the consequences or results of an event. \\
        - \textit{``What are the negative consequences for the services if they do not evaluate their programs?''} \\
        - \textit{``In the US, what is the benefit of having a red left-turn arrow?''} \\
        - \textit{``What would happen if employers violate the legislation?''} \\
        - \textit{``What if the Hokey Pokey is really what it's all about?''} \\
        
        \hline
        \textsc{\textbf{Procedural}}:
        Asking for the procedures, tools, or methods by which a certain outcome is achieved. \\
        - \textit{``Why YM 7.5 BETA always stupidly shows me available, although I initially set it to invisible?''} \\
        - \textit{``How did the Amish resist assimilation into the current social status in the U.S?''} \\
        - \textit{``How astronomers detect a nebula when there are no stars illuminating it?''} \\
        
        \hline
        \textsc{\textbf{Judgmental}}:
        Asking for the opinions of the answerer’s own. \\
        - \textit{``Do you think that it’s acceptable to call off work for a dying-dead pet?''} \\
        - \textit{``Should I date a guy that has an identical twin?''} \\
        - \textit{``How old is too old for a guy to still live with his mother?''} \\
        \hline
    
    \end{tabular}
    \caption{Question type annotation guideline.}
    \label{fig:ques_type_anno_guideline}
\end{table*}

\end{document}